\documentclass{article}
\usepackage{ijcai25}

\usepackage{times}
\usepackage{soul}
\usepackage{url}
\usepackage[hidelinks]{hyperref}
\usepackage[utf8]{inputenc}
\usepackage[small]{caption}
\usepackage{graphicx}
\usepackage{amsmath}
\usepackage{amsthm}
\usepackage{booktabs}
\usepackage{algorithm}
\usepackage{algorithmic}
\usepackage[switch]{lineno}

\usepackage[table]{xcolor}
\usepackage{multirow}
\colorlet{tableColor}{gray!20}
\usepackage{booktabs} 
\usepackage{soul}
\usepackage{xcolor,colortbl} 
\usepackage{changepage,threeparttable} 

\title{Text-driven Motion Generation: Overview, Challenges and Directions}

\author{
    Ali Rida Sahili$^1$ \and Najett Neji$^1$ \and Hedi Tabia$^1$
    \affiliations
    $^1$IBISC, Univ. Evry, Université Paris-Saclay
    \emails
    ali-rida.sahili@universite-paris-saclay.fr, najett.neji@univ-evry.fr, hedi.tabia@univ-evry.fr 
}

\begin{document}

\maketitle

\begin{abstract}    
    Text-driven motion generation offers a powerful and intuitive way to create human movements directly from natural language. By removing the need for predefined motion inputs, it provides a flexible and accessible approach to controlling animated characters. This makes it especially useful in areas like virtual reality, gaming, human-computer interaction, and robotics. In this review, we first revisit the traditional perspective on motion synthesis, where models focused on predicting future poses from observed initial sequences, often conditioned on action labels. We then provide a comprehensive and structured survey of modern text-to-motion generation approaches, categorizing them from two complementary perspectives: (i) architectural, dividing methods into VAE-based, diffusion-based, and hybrid models; and (ii) motion representation, distinguishing between discrete and continuous motion generation strategies. In addition, we explore the most widely used datasets, evaluation methods, and recent benchmarks that have shaped progress in this area. With this survey, we aim to capture where the field currently stands, bring attention to its key challenges and limitations, and highlight promising directions for future exploration. We hope this work offers a valuable starting point for researchers and practitioners working to push the boundaries of language-driven human motion synthesis.
\end{abstract}

\begin{table*}[t]
\begin{adjustwidth}{-2.5 cm}{-2.5 cm}
\centering

\setlength{\arrayrulewidth}{0.5mm} 
\setlength{\tabcolsep}{3.5pt}        
\renewcommand{\arraystretch}{0.90}  

\begin{threeparttable}[!htb]
\caption{Table summarizes the advantages and limitations of each category and the common limitations over all categories.}\label{tab:all-categories-cons-pros}
\scriptsize
\begin{tabular}{p{1.75cm}p{5.75cm}p{5.5cm}p{4.5cm}}\toprule
\rowcolor{gray!40} & & & \\
\rowcolor{gray!40} \multicolumn{1}{c}{\multirow{-2}{*}{\textbf{Category}}} &\multicolumn{1}{c}{\multirow{-2}{*}{\textbf{Advantages}}} &\multicolumn{1}{c}{\multirow{-2}{*}{\textbf{Limitations}}}  &\multicolumn{1}{c}{\multirow{-2}{*}{\textbf{Shared Limitations}}}\\
\midrule
\rowcolor{tableColor} &– Diversity &– might present some unrealistic results & \cellcolor[gray]{0.88} \\
\rowcolor{tableColor} &– Interpretability &  \quad (fails to meet physics laws) & \cellcolor[gray]{0.88} \\ 
\rowcolor{tableColor} \multirow{-3}{*}{VAE-based} &– Continuous / Discrete Latent Space &– Lack of generalization (i.e. Mode Collapse) &{\cellcolor[gray]{0.88} – Lack of diverse and natural datasets} \\
\rowcolor{gray!30} &– A promising motion learner and interpreter &– Data Dependency &{\cellcolor[gray]{0.88} (leading to lack of generalization)} \\
\rowcolor{gray!30} &– Suitable for models conditioned on multiple modalities &– Lack of Interpretability & \cellcolor[gray]{0.82} \\
\rowcolor{gray!30} \multirow{-3}{*}{Diffusion-based} &– Controllability &– Computational Complexity &{\cellcolor[gray]{0.82}– Faces and hand fingers are not fully} \\
\rowcolor{tableColor} &– Enhanced Quality &– Limited to capture fine-grained details & {\cellcolor[gray]{0.82}studied (limited for body motions)}  \\
\rowcolor{tableColor} \multirow{-2}{*}{Hybrid} &– Reduced Computational Cost &– Computational complexity is still high  & \cellcolor[gray]{0.82} \\

\bottomrule
\end{tabular}
\end{threeparttable}
\end{adjustwidth}
\end{table*}

\section{Introduction}\label{sec:intro}

    \paragraph{} Human motion generation from text leverages the natural expressiveness of language to specify complex actions. Language provides an intuitive, high-level interface for animation: a user can describe desired motion in ordinary words rather than designing poses or trajectories by hand. Unlike motion-based or physics-based systems, text-driven methods typically require no seed pose or motion clip, enabling generation from scratch purely from a description. Recent breakthroughs in deep generative modeling have made this feasible: autoregressive models, VAEs, GANs, and diffusion models have succeeded in domains as diverse as text, images, video and 3D objects~\cite{zhu2023human}, suggesting similar gains are possible for motion. Likewise, advances in large language models (LLMs) have greatly improved the ability to interpret and ground textual semantics~\cite{zhu2023human,Wang2024QuoVM}. Together, these trends motivate the study of text-to-motion generation: generating realistic human motion sequences from unrestricted natural language. \\

    Text is a flexible and expressive modality for specifying motion. A short sentence or paragraph can capture complex intent (e.g. “walk forward briskly, then wave both hands”), and does not require the user to sketch an initial posture or trajectory. Unlike future-motion prediction or motion-in-between tasks, text-to-motion models generally generate entire sequences anew given only a text prompt, rather than extending from a seed motion. This greatly broadens applicability (e.g. story-driven animation) since no pre-captured motion is needed. Modern deep-learning architectures and LLMs make it possible to translate linguistic content into other modalities. Inspired by the success of LLMs, the motion generation community is moving toward “large motion models” that can generalize to diverse actions~\cite{Wang2024QuoVM}. For example, Zhu et al. note that recent generative methods (RNNs, VAEs, flows, GANs, diffusion) have achieved great success across language and vision domains~\cite{zhu2023human}. These techniques now underpin text-to-motion systems. In addition, generating human motion has long been important in graphics, games, virtual reality, and robotics (e.g. character animation, telepresence, motion planning). Text-driven generation promises to streamline content creation and human-robot interaction by allowing designers and users to “talk” to motion models. As Guo et al. observe, translating text descriptions into human motion is a rapidly emerging topic~\cite{guo2022generating}. \\

    \paragraph{} Despite these opportunities, text-conditioned motion generation remains challenging: human motion is highly nonlinear and articulated, and mapping ambiguous language to realistic dynamics is complex~\cite{zhu2023human}. As Zhu et al. point out, even with recent advances, the task poses difficulties due to the intricate nature of motion and its implicit relationship with language context~\cite{zhu2023human}. This survey seeks to clarify the state of the art and outline directions for overcoming these challenges. \\

    \paragraph{} Several recent surveys have reviewed human motion modeling, but none focus exclusively on text-driven generation. Notably, Mourot et al.~\cite{mourot2022survey} present a comprehensive survey of deep-learning methods for skeleton-based character animation. They cover motion synthesis, character control, and editing under a range of deep models, but do not address language as a conditioning modality. Lyu et al.~\cite{lyu20223d} survey “3D human motion prediction,” which concerns forecasting future poses given past frames. That task is closely related to animation but fundamentally different: it conditions on motion history rather than textual commands. Zhu et al.~\cite{zhu2023human} offer the first broad survey of human motion generation, reviewing methods conditioned on text, audio, and scene. While Zhu does include text-conditioned generation among its subtopics, the scope remains multi-modal and general. Their survey covers various datasets and metrics across all modalities.

    In contrast, this review hones in specifically on text-to-motion generation. We delve deeper into methods that interpret natural language and translate it to human motion. Unlike the works by Mourot et al.\cite{mourot2022survey} and Lyu et al.\cite{lyu20223d}, which do not address the language aspect of motion generation, and in contrast to Zhu et al.~\cite{zhu2023human}, who consider text as just one of several conditioning modalities, our survey offers a more focused and up-to-date exploration of text-driven motion generation methods.
        
    \paragraph{} In this review, we offer a comprehensive and focused overview of recent advances in text-to-motion generation. To the best of our knowledge, this is the first survey to systematically analyze the field from two complementary perspectives: (1) architectural paradigms, where we classify methods into VAE-based, diffusion-based, and hybrid approaches—each representing distinct generative strategies and modeling philosophies; and (2) motion representation spaces, where we distinguish between discrete and continuous frameworks, depending on how human motion is structured and synthesized. Alongside this dual categorization, we provide a detailed comparison of available language-motion datasets, a critical examination of commonly used evaluation metrics, and an analysis of key challenges and future research directions. Our aim is to consolidate current knowledge in this fast-moving area and to offer a valuable reference for both newcomers and researchers working on language-driven human motion synthesis.

\section{Problem Scope and Early Methods}\label{sec:traditional-motion-generation}

    \subsection{Problem Formulation}
    
    \paragraph{} Formally, we define text-to-motion generation as the task of learning a function or distribution that maps a textual description to a time series of human poses. Let a text prompt $T$ be defined as a sequence of words or tokens $(w_1, w_2, \dots, w_L)$ that describes an action (e.g. “a person jumps forward three times”). We seek to generate a corresponding motion sequence $\mathcal{M} = \{\mathbf{p}_1, \mathbf{p}_2, \dots, \mathbf{p}_N\}$, where each $\mathbf{p}_t$ represents the human pose at time step t. The representation of each pose $\mathbf{p}_t$ can vary depending on the model and dataset used, as discussed in the following paragraph. The goal is to learn a model $G$ such that $\mathcal{M} = G(T)$ produces a realistic, temporally coherent motion that semantically matches $T$. Equivalently, models often learn the conditional distribution $p(\mathcal{M} | T)$, enabling stochastic sampling of diverse motions for the same text. 
    
    Typical data representations for motion include: joint positions presented as 3D coordinates of body joints per frame or 2D projections on the image plane, joint rotations expressed as Euler angles or axis-angle parameters for skeletal joints, and parametric body models (e.g. SMPL) that define full-body mesh shape and pose. These representations can be interconverted (e.g. joint angles to positions via forward kinematics). In this survey, we consider any of these forms as valid outputs of $\mathcal{M}$. As noted by Zhu et al.~\cite{zhu2023human}, human motion is often encoded as sequences of 2D/3D keypoints, joint rotations, or parametric model parameters. 

    \paragraph{} The generation objective is to produce motion that is both realistic and relevant to the text. In practice, methods are trained to minimize reconstruction or adversarial losses on pose sequences while aligning with semantic constraints implied by $T$. Some works explicitly optimize an alignment between textual and motion embedding spaces. Overall, success is measured by whether the synthesized motion obeys natural human dynamics (balance, physics constraints, smooth transitions) and satisfies the action described by $T$.

    \subsection{Traditional Perspective}

    \paragraph{} Before exploring modern text-driven approaches to motion generation, it is important to situate them within the broader historical context of motion synthesis. Earlier methods primarily approached the task as a conditional motion prediction problem—aiming to forecast future poses based on a sequence of observed initial poses. These traditional models often relied on action labels, commonly encoded as one-hot vectors, to guide the generation process toward specific activities. While such approaches helped establish the foundation for more advanced generative techniques, they were inherently limited in flexibility and expressiveness. Their reliance on predefined motion cues and simplistic semantic inputs often restricted the diversity and richness of the generated motion.

    \paragraph{} One of the seminal works in this area was introduced by Martinez et al.~\cite{martinez2017human}, who proposed a simple yet effective Recurrent Neural Network (RNN)-based sequence-to-sequence (Seq2Seq) model with residual connections. The model took a sequence of previous poses as input and produced a forecast of future poses. To incorporate semantic information, a one-hot encoded action label was appended to the input vector. Despite its architectural simplicity, this method demonstrated competitive performance in short-term motion prediction tasks and established a strong baseline for future works.

    \paragraph{} To improve the realism and generalization of generated motions, Hernandez et al.~\cite{hernandez2019human} introduced a multi-discriminator adversarial training setup. Their architecture comprised a generator and three discriminators, each operating on different spatial and temporal features: absolute joint coordinates, relative joint offsets, and temporal differences of both. Additionally, they introduced bone-length and limb-structure losses to preserve anatomical plausibility—ensuring consistent bone lengths and natural distances between extremities such as hands, feet, and head. These domain-specific constraints contributed to more realistic and coherent motion outputs.

    \paragraph{} Wang et al.~\cite{wang2020learning} further extended adversarial frameworks by using a noise-conditioned RNN-based generator, coupled with a single discriminator to classify real versus generated motions. Like previous models, they conditioned the generation process using one-hot encoded action labels, reaffirming the trend of combining minimal semantic cues with learned temporal dynamics.

    \paragraph{} The increasing interest in capturing the structural topology of the human body led researchers to adopt Graph Convolutional Networks (GCNs)~\cite{schlichtkrull2018modeling}, which treat joints as graph nodes and bones as edges. Yu et al.~\cite{yu2020structure} proposed a structure-aware generative adversarial model that combined GCNs with self-attention mechanisms~\cite{vaswani2017attention}. Their generator sampled from a latent space combining Gaussian noise and action labels, then decoded the samples using an LSTM-based architecture enhanced by GCN layers. Two discriminators were employed: one on entire motion sequences and the other on temporally sparse frame subsets, both conditioned on action labels. While this method yielded promising results in preserving skeletal structure and capturing long-term dependencies, it incurred significant computational costs and occasionally produced physically implausible motions due to overfitting or excessive regularization.

    \paragraph{} In summary, traditional approaches to motion generation primarily relied on sequence modeling techniques like RNNs and LSTMs, with minimal semantic conditioning in the form of action labels. Though effective for short-term predictions and simple activity patterns, these methods struggled with generating diverse, context-rich, and naturalistic motions. Their reliance on observed initial poses restricted their applicability to scenarios where such inputs are unavailable. The growing capabilities of natural language understanding, combined with the emergence of powerful generative models like VAEs, GANs, and diffusion models, have fundamentally transformed the way we approach motion synthesis. Instead of relying on deterministic, pose-based forecasting, we can now generate human motion directly from language turning simple text descriptions into dynamic animations. This shift represents more than just a technical milestone, it opens up new possibilities for creating expressive, scalable, and interactive motion systems that can be controlled as easily as we speak.

\section{Categorization}\label{sec:text-to-motion-categories}

\subsection{Architectural-based}
    \paragraph{} As text-driven human motion generation continues to advance, a clear pattern has emerged in the design of modern methods. Most state-of-the-art models fall into one of three main architectural categories: VAE-based, diffusion-based, and hybrid approaches. VAE-based methods typically start by learning a compact, latent representation of motion using a variational autoencoder (VAE)\cite{petrovich2022temos}. On the other hand, diffusion-based methods rely on denoising diffusion probabilistic models (DDPMs)\cite{tevet2022motionclip,zhang2022motiondiffuse}. Their ability to capture rich temporal dynamics has made them a popular choice in recent work. More recently, hybrid methods have started to gain traction. These approaches blend the strengths of both VAEs and diffusion models, often through frameworks like latent diffusion models (LDMs)~\cite{rombach2022high}. By combining the structured representation learning of VAEs with the powerful generative capabilities of diffusion models, hybrid methods aim to produce more flexible and high-quality motion outputs. This three-part categorization offers a helpful way to understand how different architectural strategies tackle the unique challenges of generating human motion from language. 
    
    Table~\ref{tab:all-categories-cons-pros} offers a bird’s-eye view of the main strengths and limitations of the three major architectural families—VAE-based, diffusion-based, and hybrid models while also highlighting some of the common challenges that cut across all categories. Tables~\ref{tab:vae-based-category},~\ref{tab:diffusion-based-category} and~\ref{tab:hybrid-category} then dive deeper, summarizing the different methods in each category and outlining their key contributions and challenges.

\subsubsection{VAE-based Methods}\label{sec:vae-based}
    \paragraph{} Among the different generative approaches for human motion synthesis, Variational Autoencoders (VAEs)~\cite{kingma2013auto} have proven to be particularly effective in modeling the underlying probabilistic structure of motion data. These models are well-suited for generating diverse, high-quality motion sequences from natural language descriptions. The core objective in this context is to model the conditional probability distribution $p(M|T)$, where $M$ denotes a motion sequence and $T$ represents a text input. Due to the inherent complexity of this distribution, VAEs reformulate the objective via the introduction of a latent variable $z$, such that:
    \begin{equation}
        P(M|T) = \int p(M|z,T)P(z|T)dz
    \end{equation} 
    Here, $z$ encapsulates a high-level representation of motion, enabling structured sampling and generation. The typical architecture includes a motion encoder, a motion decoder, and a text encoder. The encoder-decoder training is designed to minimize reconstruction loss while ensuring that the approximate posterior $q(z|M,T)$ closely follows the true posterior.

    \paragraph{} The earliest work in this domain, Action2Motion~\cite{guo2020action2motion}, leveraged an RNN-based VAE to generate 3D motion from action labels. They introduced a Lie algebra representation to model joint rotations and iteratively generated poses using autoregressive decoding. However, its performance is limited to seen actions and lacks the ability to compose complex or novel motions. Petrovich et al.~\cite{petrovich2021action} advanced this architecture by incorporating a Transformer-based conditional VAE that operates on full motion sequences and includes duration encoding through positional embeddings. Lee et al.~\cite{lee2023multiact} extended this model to handle multi-action generation, enabling composition of sequential activities, though the diversity of generated motion remained constrained. Zhai et al.~\cite{zhai2023language} further evolved the framework by introducing a codebook-guided transformer VAE. They employed a simple MLP to encode text, a masked reconstruction objective over temporal frames, and discrete atomic motion codes to enhance generalization. Nonetheless, the reliance on basic text encodings limited the model's capacity to interpret complex descriptions.

    \paragraph{} To address fusion inefficiencies, Zhong et al.~\cite{zhong2022learning} proposed a Feature-wise Linear Modulation (FiLM) strategy for conditioning on action labels. Although this improved motion-text integration, the model struggled to generate previously unseen motions. The TEMOS framework~\cite{petrovich2022temos} introduced dual encoders for motion and text—using a frozen DistilBERT~\cite{sanh2019distilbert} and learned a shared latent space with an additional embedding similarity loss to improve alignment. While effective in generating realistic motion, the model was memory-intensive and struggled with long sequences and diversity. This framework was later extended in Teach~\cite{athanasiou22teach} for smoother transitions in action composition. Action-GPT~\cite{Action-GPT} represents a unique direction by integrating large language models (LLMs) into the pipeline. The method transforms action labels into descriptive prompts using GPT-3, which are then converted into fine-grained motion features. This modular pipeline is compatible with many VAE-based backbones and opens new directions for enhancing semantic grounding.

    \paragraph{} A major advancement in VAE-based text-to-motion models involves the use of discrete latent codes through vector quantization. PoseGPT~\cite{lucas2022posegpt} and MotionBERT~\cite{zhu2023motionbert} adopt a VQ-VAE~\cite{van2017neural,razavi2019generating} structure to learn a discrete motion vocabulary. These models tokenize motions and use transformers to predict the next latent token, conditioned on text and duration. This approach significantly improves the diversity and control over motion generation.

    Guo et al.~\cite{guo2022generating} proposed a two-stage framework where text determines sequence length (Text2Length) before generating motion tokens in an autoregressive fashion (Text2Motion), followed by a pretrained decoder. However, the model had difficulties with rare actions and fine-grained localization. Subsequent work~\cite{guo2022tm2t} updated this structure with GPT-like text encoders~\cite{radford2018improving} and improved motion token prediction, leading to models like T2M-GPT~\cite{zhang2023t2m} and MotionGPT~\cite{jiang2023motiongpt}, which rival diffusion-based models in quality. Nevertheless, they still face generalization limits when generating motions not seen in the training data.

    Pinyoanuntapong et al.~\cite{pinyoanuntapong2024mmm} proposed a Conditional Masked Transformer model where CLIP embeddings~\cite{radford2021learning} and discrete motion tokens are combined for generation and editing. Recent extensions~\cite{guo2024momask,pinyoanuntapong2025bamm} further refined the masking strategies but remained limited in modeling rapid root transitions, a common issue in VQ-based systems.

    \paragraph{} VAE-based methods have shown strong potential in generating coherent, high-quality, and text-conditioned human motions. Their capacity to learn structured latent spaces makes them a powerful foundation for motion synthesis. However, limitations persist in generalization, especially for complex, composite actions and underrepresented motion types. Most approaches also overlook fine details like facial expressions and finger movements, which are crucial for truly lifelike motion generation. \\

    Despite these advances, challenges such as modeling intricate multimodal distributions, improving semantic alignment with complex textual prompts, and scaling to long-term motion planning remain. These shortcomings have paved the way for diffusion-based models, which are increasingly favored for their ability to generate high-fidelity and temporally consistent motion sequences. 


\begin{table*}[htbp]
\begin{adjustwidth}{-2.5 cm}{-2.5 cm}
\centering
\begin{threeparttable}[!htb]
\caption{Summary of main contributions and limitations of VAE-based Methods for text-to-motion generation.}\label{tab:vae-based-category}
\scriptsize
\begin{tabular}{p{3.65cm}p{7.5cm}p{7.0cm}}\toprule
\multicolumn{1}{c}{\multirow{2}{*}{\textbf{Method}}} &\multicolumn{1}{c}{\multirow{2}{*}{\textbf{Main Contributions}}} &\multicolumn{1}{c}{\multirow{2}{*}{\textbf{Limitations}}} \\
& & \\\midrule
\multirow{4}{*}{Action2Motion \cite{guo2020action2motion}} &– First method in action-conditioned motion generation &– Lack of generalization \\
&– Lie Algebra based VAE framework &– limited to generate non-complex motions of one action \\
&– Building new 3D human motion dataset: HumanAct12 & \\ & & \\
\multirow{3}{*}{ACTOR \cite{petrovich2021action}} &– Transformer-based conditional VAE &– Data Dependency \\
&– A non-learnable differentiable SMPL layer &– Computational intensive to generate long sequences \\ & & \\
\multirow{4}{*}{T2M \cite{guo2022generating}} &– text2length sampling stage to determine the motion duration &– It fails in descriptions involving rare actions (e.g., ’stomp’) \\
&– text2motion stage to generate motion with a Temporal VAE &– It fails in finer grained descriptions and complex actions. \\
& &– It shows some unrealistic generated motions \\ & & \\
\multirow{3}{*}{TM2T \cite{guo2022tm2t}} &– Motion Tokenizer to transform motions into discrete codes &– some unrealistic generated motions \\
&– Inverse Alignment Technique &– Not able to deal with long and complex descriptions \\ & & \\
\multirow{3}{*}{T2M-GPT \cite{zhang2023t2m}} &– VQ-VAE to transform motions into discrete codes &– Not able to generate novel motions not seen in the training data \\
&– GPT-like model to encode textual descriptions & \\ & & \\
\multirow{3}{*}{MotionGPT \cite{jiang2023motiongpt}} &– Motion Tokenizer to transform motions into discrete codes &– Not designed for hands and faces motion \\
&– T5 model to get text embedding &– Lack of generalization \\ & & \\
\multirow{4}{*}{TEMOS \cite{petrovich2022temos}} &– non-autoregressive transformer-based VAE &– Quadratic memory cost (not suitable for long motions) \\
&– A DistilBERT model to encode text descriptions &– it might drastically fail if the input text contains typos \\
&– Cross-modal Embedding Similarity &– lack of diversity (limited diversity) \\ & & \\
\multirow{3}{*}{TEACH \cite{athanasiou22teach}} &– Benchmark for temporal action composition of 3D motions &– susceptible to acceleration peaks between actions transitions \\
&– Extends TEMOS to be able to handle series of actions & \\ & & \\
\multirow{3}{*}{ImplicitMotion \cite{cervantes2022implicit}} &– Variational Implicit Neural Representation &– Some poor performance depending on parameters updates \\
& &– linear computational cost \\ & & \\
\multirow{3}{*}{ATOM \cite{zhai2023language}} &– CVAE (complex actions decomposed into atomic actions) &– limited ability to interpret complex text descriptions \\
&– Curriculum Learning strategy based on masked motion modeling &– limited fusion strategy between text and motion features \\ & & \\
\multirow{3}{*}{UM-CVAE \cite{zhong2022learning}} &– Uncoupled sequence-level CVAE &– Unable to generate entirely new and unseen actions \\
&– FiLM-based action-aware modulation &– Limited quality of generated motions (data dependency) \\ & & \\
\multirow{3}{*}{PoseGPT \cite{lucas2022posegpt}} &– GPT-like model for next-index prediction in the latent space &– the quantization scheme might limit the model to generate \\
&– Encoder-Decoder architecture eith quantized latent space & diverse motions \\ & & \\
\multirow{3}{*}{MultiAct \cite{lee2023multiact}} &– conditional VAE Architecture &– Generation of some unrealistic motions \\
&– Model to generate long 3d motion sequence of multiple actions &– Falied to generate complex and diverse action sequences \\ & & \\
\multirow{4}{*}{Action-GPT \cite{Action-GPT}} &– First LLMs-based text-conditioned motion generation approach &– Not able to generate long sequences \\
&– A module compatible with VAE-based models &– Cannot generate complex body movement (e.g. yoga and dance) \\
& &– Finger motion is not supported \\ & & \\
\multirow{3}{*}{TMR \cite{petrovich2023tmr}} &– Contrastive learning over VAE latent space &– lack of generalization \\
&– Filtering strategy of similarities in textual descriptions &– memory-inefficient at some use cases \\ & & \\
\multirow{3}{*}{TM2D \cite{gong2023tm2d}} &– VQ-
VAE framework   &– lack of paired data (music, text)   \\
&– Bimodal feature fusion strategy (Cross-Modal Transformer) &– limited to specific dance styles and genres (data dependency)  \\ & & \\
\multirow{3}{*}{T2LM \cite{lee2024t2lm}} &– Continuous long-term VQ-VAE generation framework  &– Limited to generate fine-grained motions  \\
&– 1D-convolutional VQVAE (to avoid temporal inconsistencies)  &– Limited to short textual descriptions  \\ & & \\
\multirow{3}{*}{AttT2M \cite{zhong2023attt2m}} &– Body-Part attention-based Spatio-Temporal VQ-VAE  &– Insufficient diversity in long and detailed text-driven generation  \\
&– Global and Local Attention to learn the cross-modal relationship &– Data Dependency (unable to generate unseen motions) \\ & & \\
\multirow{2}{*}{MMM \cite{pinyoanuntapong2024mmm}} &– Conditional Masked Motion Model &– Insufficient to generate motion for long and detailed textual descriptions  \\ & & \\
\multirow{3}{*}{MoMask \cite{guo2024momask}} &– Motion Residual VQ-VAE &– Limited diversity \\
&– Conditional Masked Transformer &– Fails to generate motions with fast-changing root motions \\ & & \\
\multirow{2}{*}{BAMM \cite{pinyoanuntapong2025bamm}} &– Conditional Masked Self-attention Transformer &– moderate computational complexity \\
&– Hybrid Attention Masking for Training &– Fails to generate motions with fast-changing root motions \\
\bottomrule
\end{tabular}
\end{threeparttable}
\end{adjustwidth}
\end{table*}

\begin{table*}[htbp]
\begin{adjustwidth}{-2.5 cm}{-2.5 cm}
\centering
\begin{threeparttable}[!htb]
\caption{Summary of main contributions and limitations of diffusion-based Methods for text-to-motion generation.}\label{tab:diffusion-based-category}
\scriptsize
\begin{tabular}{p{3.5cm}p{8cm}p{6.25cm}}\toprule
\multicolumn{1}{c}{\multirow{2}{*}{\textbf{Method}}} &\multicolumn{1}{c}{\multirow{2}{*}{\textbf{Main Contributions}}} &\multicolumn{1}{c}{\multirow{2}{*}{\textbf{Limitations}}} \\
& & \\\midrule
\multirow{4}{*}{MotionDiffuse \cite{zhang2022motiondiffuse}} &– First diffusion model-based text-driven motion generation framework. &– It requires a large amount of diffusion steps during inference \\
&– It utilizes a DDPM model &– It shows some unrealistic generated motions. \\
&– a Cross-Modality Linear Transformer to process input sequences & \\ & & \\
\multirow{3}{*}{MDM \cite{tevet2022human}} &– Transformer-based diffusion model &– high computational overheads and low inference speed \\
&– classifier free guidance for the diffusion process &– suitable for short motion sequence generation \\ & & \\
MMDM \cite{chen2024text} &– Mask Modeling Strategy over time frames and body parts &– Computationally Expensive \\ 
\multirow{5}{*}{priorMDM \cite{shafir2023human}} & & \\ &– parallel composition: two single motions performing together &– limited to the quality of the initial model \\
&– sequential composition: long animations with different actions &– motion inconsistencies between distant intervals \\
& &– lacking generalization \\ & & \\
\multirow{4}{*}{FlowMDM \cite{barquero2024seamless}} &– Blended Positional Encodings &– Fails for complex text descriptions \\
&– Pose-centric cross-attention &– Slight mismatch for some transitions \\
&– Building two new metrics (to detect sudden transitions) & \\ & & \\
\multirow{3}{*}{Physdiff \cite{yuan2023physdiff}} &– Injecting physical constraints in the generation process &– High Computational Complexity of incorporating a \\
&– Joint training framework to learn from both real motion data and simulated data & physics simulator in the diffusion process \\ & & \\
\multirow{3}{*}{MoFusion \cite{dabral2023mofusion}} &– lightweight 1D U-Net network with cross-modal transformers for reverse diffusion &– High inference time \\
&– time-varying weight schedule for Kinematic losses &– restricted vocabulary for textual conditioning \\ & & \\
\multirow{3}{*}{FLAME \cite{kim2023flame}} &– Pre-trained large model for text encoding (Roberta) &– Computationally Expensive \\
&– masking strategy on trasnformer decoder & \\ & & \\
\multirow{3}{*}{GMD \cite{karunratanakul2023gmd}} &– Incorporating Spatial Constraints (predefined trajectories, obstacles avoidance) &– Data Dependency \\
&– Feature Projection Scheme &– Computationally Expensive \\ & & \\
\multirow{3}{*}{DiffGesture \cite{zhu2023taming}} &– Diffusion Audio-Gesture Transformer (attend information from mutli-modalities) &– Limited Data Diversity \\
&– Diffusion Gesture Stabilizer to eliminate temporal inconsistency &– Computationally Expensive \\ & & \\
\multirow{3}{*}{LDA \cite{alexanderson2023listen}} &– Conformer-based diffusion model &– Speech Feature Extraction Dependency \\
&– Built a new dataset with audio and high-quality 3D motion capture &– Computationally Expensive \\ & & \\
\multirow{3}{*}{ReMoDiffuse \cite{zhang2023remodiffuse}} &– retrieval-augmented motion diffusion model &– Data Dependency \\
&– Semantics-modulated transformer &– Computationally Expensive \\ & & \\
\multirow{3}{*}{FineMoGen \cite{zhang2023finemogen}} &– Spatio-Temporal Mixture Attention &– Limited to various motion data formats \\
&– Building large-scale language-motion dataset (HuMMan-MoGen) &– LLM dependency \\ & & \\
\multirow{3}{*}{Fg-T2M \cite{wang2023fg}} &– Linguistics-Structure Assisted Module &– Limited to the capabilities of the used language model \\
&– Context-Aware Progressive Reasoning Module & \\ & & \\
\multirow{3}{*}{MAA \cite{azadi2023make}} &– Pre-training the diffusion model on text-conditioned 3d static poses dataset &– High Computational Cost \\
&– Temporal Extension and Fine-tuning on text-driven motion generation dataset &– generates some  unnatural movements \\ & & \\
\multirow{3}{*}{StableMoFusion \cite{huang2024stablemofusion}} &– Effective mechanism to eliminate foot skating &– Slow Inference (High Computational Cost) \\
&– Systematic analysis of different components of the diffusion-based motion generation pipeline & \\ & & \\
\bottomrule
\end{tabular}
\end{threeparttable}
\end{adjustwidth}
\end{table*}

\begin{table*}[t]
\begin{adjustwidth}{-2.5 cm}{-2.5 cm}
\centering
\begin{threeparttable}[!htb]
\caption{Summary of main contributions and limitations of Hybrid Methods for text-to-motion generation.}\label{tab:hybrid-category}
\scriptsize
\begin{tabular}{p{3.5cm}p{6.25cm}p{5.75cm}}\toprule
\multicolumn{1}{c}{\multirow{2}{*}{\textbf{Method}}} &\multicolumn{1}{c}{\multirow{2}{*}{\textbf{Main Contributions}}} &\multicolumn{1}{c}{\multirow{2}{*}{\textbf{Limitations}}} \\
& & \\\midrule
\multirow{3}{*}{MLD \cite{chen2023executing}} &– A motion transformer-based VAE &– Limited generated motion length \\
&– Conditional diffusion at latent space &– Limited to human bodies (no hands or faces) \\ & & \\
\multirow{3}{*}{M2DM \cite{kong2023priority}} &– Transformer-based VQ-VAE &– Limited to capture fine-grained details in the motion \\
&– priority-centric scheme & \\ & & \\
MoDDM \cite{Chemburkar_2023_BMVC} &– a discrete denoising diffusion probabilistic model &– Limited to capture fine-grained details in the motion \\
\multirow{4}{*}{UDE \cite{zhou2023ude}}  & & \\ &– Single text-driven and audio-driven motion generation model &– struggle to handle complex interactions between different modalities \\
&– Diffusion-based Decoder & \\  & & \\
\multirow{3}{*}{GestureDiffuCLIP \cite{ao2023gesturediffuclip}} &– style control by multimodal prompts (text and speech) &– Data Dependency \\
&– CLIP-guided prompt-conditioned co-speech gesture synthesis &– CLIP Limitations for detailed motions \\ & & \\
\multirow{3}{*}{M2D2M \cite{chi2024m2d2m}} &– Dynamic transition probability model & \\
&– New evaluation metric (Jerk) to assess the smoothness at action boundaries &– Jerk is not able to assess all generated scenarios \\ & & \\
\multirow{3}{*}{EMDM \cite{zhou2025emdm}} &– A conditional denoising diffusion GAN &– Physical laws might not be respected (e.g. floating, \\
&– Faster diffusion-based approach & ground penetration)  \\ & & \\
\multirow{2}{*}{Motion Mamba \cite{zhang2025motion}} &– Denoising U-Net with two modules: Hierarchical  &– Performance for short sequences is not shown \\
&Temporal Mamba and Bidirectional Spatial Mamba &– model's generalization is not studied \\
\bottomrule
\end{tabular}
\end{threeparttable}
\end{adjustwidth}
\end{table*}

\subsubsection{Diffusion-based Methods}\label{sec:diffusion-based}
    \paragraph{} Building upon the limitations observed in VAE-based methods—particularly in generating diverse, temporally coherent, and high-fidelity motion sequences—diffusion models have emerged as a powerful alternative for text-to-motion generation. Originally proposed for image synthesis~\cite{ho2020denoising}, denoising diffusion probabilistic models (DDPMs) have been successfully adapted to motion generation due to their robust ability to model complex, multimodal distributions in high-dimensional spaces.

    At a high level, diffusion models learn to reverse a gradual noising process that transforms data into pure noise. Specifically, a sequence of motion data is corrupted over multiple timesteps through the addition of Gaussian noise. The model is then trained to denoise this corrupted data in a step-wise manner, gradually recovering the original motion sequence. For conditional generation, the denoising network is guided by a conditioning signal—in this case, a text description—to produce motion sequences that align with the semantic content of the input.

   \paragraph{} The pioneering work of MotionDiffuse~\cite{zhang2022motiondiffuse} introduced this approach to human motion synthesis. It adopts a denoising autoencoder trained on human motion sequences, with CLIP-based [3] text embeddings serving as conditioning input. The motion is represented in joint angle format, and the model uses a transformer-based architecture to model temporal dependencies. Despite its simplicity, MotionDiffuse achieves high-quality and temporally consistent motion, outperforming many VAE-based baselines on standard benchmarks.

   Subsequent work has focused on improving motion representation, conditioning mechanisms, and sampling efficiency. In MDM (Motion Diffusion Model)~\cite{tevet2022human}, Tevet et al. propose a transformer-based denoising network with a learned motion representation. They leverage pre-trained CLIP embeddings for text guidance and design a time-aware attention mechanism to improve long-range consistency. MDM set a new state-of-the-art in text-conditioned motion generation and laid the groundwork for several follow-up models. Chen et al.~\cite{chen2024text} extended this model by introducing a spatio-temporal masking strategy, encouraging better generalization across body parts and time. Shafir et al.~\cite{shafir2023human} built on this by generating long-form and multi-person interactions. Barquero et al.~\cite{barquero2024seamless} proposed a novel blending of positional encodings to achieve smooth transitions when switching between motions. \\

    MoFusion~\cite{dabral2023mofusion} focused on efficient architecture design, introducing a lightweight 1D U-Net paired with three kinematic consistency losses. This model significantly improved long-term coherence and reduced inference latency. Kim et al.~\cite{kim2023flame} proposed FLAME, a transformer decoder-only model with dynamic masking that flexibly handles variable-length inputs. Zhang et al.~\cite{zhang2023remodiffuse} addressed retrieval-augmented generation via ReMoDiffuse, which integrates top-k similar motion samples and semantic features from CLIP to guide the generative process. To improve motion diversity and realism, Azadi et al.~\cite{azadi2023make} introduced Make-an-Athlete, which utilizes a large-scale pseudo-pose dataset derived from image-text corpora. The model is trained in two phases: first on static pose generation, and then temporally extended and fine-tuned on real motion data. This resulted in state-of-the-art performance on diversity metrics.

    \paragraph{} Despite their strengths, diffusion models are computationally intensive due to the iterative denoising process applied directly in motion space. Their performance is also sensitive to the training dataset's diversity, reaffirming the need for broader and richer motion-text corpora.

\subsubsection{Hybrid Methods}\label{sec:hybrid-methods}
    \paragraph{} Beyond the dichotomy of VAEs and diffusion models, recent research has introduced a hybrid class that combines the strengths of both paradigms—particularly the structured latent representations of VAEs and the expressive generative modeling of diffusion networks. These hybrid methods typically apply diffusion in a learned latent space rather than the raw motion space, drastically reducing computational demands and improving generative quality.

    \paragraph{} One foundational method is Latent Diffusion~\cite{rombach2022high}, initially introduced in the image domain, which applies diffusion in the latent space of a pretrained VAE. Chen et al.~\cite{chen2023executing} extended this idea to motion synthesis by training a VAE on motion sequences and then applying a denoising diffusion model to its latent representations. This two-stage pipeline significantly improves sample efficiency and motion fidelity, outperforming earlier models such as MDM. Kong et al.~\cite{kong2023priority} pushed this direction further with a discrete latent representation. Their approach tokenizes motion into discrete units and employs a mask-and-replace strategy during training, selectively corrupting tokens to enable robust reconstruction. This method supports both generation and editing tasks with enhanced flexibility.

    M2D2M~\cite{chi2024m2d2m}, proposed by Chi et al., builds upon vector quantization techniques to map motions into discrete token spaces. A transformer-based denoising model then reconstructs motion sequences from corrupted tokens, conditioned on CLIP text embeddings. This design allows for efficient training and high-quality synthesis of multi-action motion sequences. Zhou et al.~\cite{zhou2025emdm} introduced EMDM, a novel combination of conditional diffusion and adversarial training. This model employs a diffusion generator that incorporates latent noise, text features, and partially observed motion data. A conditional discriminator then evaluates generated samples based on alignment with the text prompt. While powerful, EMDM's use of diffusion in raw motion space leads to high computational costs and occasional violations of physical constraints.\\

    \paragraph{} Hybrid models, therefore, present a promising direction that bridges the scalability of VAEs and the fidelity of diffusion models. By operating in latent or tokenized spaces, they enable more efficient training and inference, while maintaining diversity and realism. However, they still face limitations in capturing fine-grained motion details, particularly in underrepresented scenarios.  \\   

    Despite the substantial progress made in text-to-motion generation, two critical challenges persist across all model categories. First is the limited availability of large-scale, diverse, and well-annotated datasets. Without sufficient training data, even the most advanced models struggle to generalize to unseen or complex queries. Second is the narrow focus on full-body motions, often neglecting detailed articulation such as facial expressions and finger movements. Future research should emphasize richer data collection and fine-grained modeling to bridge these gaps.


\subsection{Motion-Space-based}

    \paragraph{} one of the most fundamental design choices lies in how motion data is represented: via discrete or continuous latent spaces. This decision critically shapes the architecture, performance, and flexibility of generative models. Both paradigms aim to encode a motion sequence $M$ into a latent representation $z=f(M)$, fuse this with textual guidance, and decode the result into coherent motion. However, they diverge significantly in methodology and downstream capabilities. Discrete approaches typically quantize motions into a finite “vocabulary” of tokens (often via a VQ-VAE), enabling the use of language-modeling techniques. By contrast, continuous methods model motions directly (or via a continuous latent) and often employ diffusion or similar generative processes. \\


\begin{table*}[t]
\begin{adjustwidth}{-2.5 cm}{-2.5 cm}
\centering

\setlength{\arrayrulewidth}{0.5mm} 
\setlength{\tabcolsep}{3.5pt}        
\renewcommand{\arraystretch}{1.60}  

\begin{threeparttable}[!htb]
\caption{Comparison of Discrete vs. Continuous Representations in Text-to-Motion Generation.}\label{tab:motion-space-cat}
\scriptsize
\begin{tabular}{p{4.5cm}p{5.5cm}p{5.5cm}}\toprule
\multicolumn{1}{c}{\multirow{-1}{*}{\textbf{Aspect}}} &\multicolumn{1}{c}{\multirow{-1}{*}{\textbf{Discrete Representations}}} &\multicolumn{1}{c}{\multirow{-1}{*}{\textbf{Continuous Representations}}}\\
\midrule

\textbf{Motion Encoding} & VQ-VAE or quantizer produces motion tokens from pose sequences & Autoencoders or raw continuous pose data directly used \\
\textbf{Generative Models} & Transformers (e.g., GPT), Masked models (e.g., BERT), Discrete diffusion & Diffusion models on raw motion or latent spaces (e.g., LDMs) \\
\textbf{Alignment with Text} & Easy to integrate with NLP models; can treat motion as a language & Requires attention/cross-modal fusion; less structured mapping \\
\textbf{Training Stability} & Prone to codebook collapse and quantization artifacts & Generally stable due to continuous MSE losses in diffusion \\
\textbf{Fidelity and Diversity} & High fidelity if codebook is large enough; limited diversity & Naturally diverse and expressive via stochastic sampling \\
\textbf{Inference Speed} & Faster for small autoregressive models; slower for long sequences & Generally slow due to iterative sampling; speed improved with LDMs \\
\textbf{Control and Editing} & Enables inpainting and symbolic-level control via tokens & Fine-grained editing (e.g., FLAME, SALAD) via frame/joint control \\
\textbf{Streaming/Online Capability} & Limited by autoregressive decoding; non-causal & Easier to support streaming generation with causal latents (e.g., MotionStreamer) \\
\textbf{Common Limitations} & Information loss from quantization; difficulty in training tokenizer & High computational cost; harder to align precisely with text \\
\textbf{Representative Works} & T2M-GPT~\cite{zhang2023t2m}, MMM~\cite{pinyoanuntapong2024mmm}, MotionGPT~\cite{jiang2023motiongpt}, MoDDM~\cite{Chemburkar_2023_BMVC}, M2D2M~\cite{chi2024m2d2m} & MotionDiffuse~\cite{zhang2022motiondiffuse}, MoFusion~\cite{dabral2023mofusion}, FLAME~\cite{kim2023flame}, SALAD~\cite{hong2025salad}, MoLA~\cite{uchida2024mola}, MotionStreamer~\cite{xiao2025motionstreamer} \\

\bottomrule
\end{tabular}
\end{threeparttable}
\end{adjustwidth}
\end{table*}

    \subsubsection{Discrete latent representations}
        \paragraph{} Discrete methods first encode motion sequences into a series of token indices using a learned codebook, and then model token sequences with generative models (e.g. Transformers). A typical pipeline uses a VQ-VAE or vector-quantized autoencoder~\cite{van2017neural} as a motion tokenizer. For instance, MMM (Masked Motion Model)~\cite{pinyoanuntapong2024mmm} employs a VQ-VAE that learns a large codebook (8192 entries) to quantize 3D motion embeddings. This discrete tokenizer converts and quantizes raw motion data into a sequence of discrete motion tokens~\cite{pinyoanuntapong2024mmm}, preserving fine-grained motion details via high-resolution quantization. Once tokenized, various generative stages are possible. \\
        
        In T2M-GPT~\cite{zhang2023t2m} and MotionGPT~\cite{jiang2023motiongpt}, a Transformer-based language model is trained to predict motion tokens conditioned on text. T2M-GPT uses a CNN+VQ-VAE followed by a GPT with a corruption strategy, and finds that this simple two-stage pipeline shows better performance than competitive approaches, including recent diffusion-based approaches. Similarly, MotionGPT uses discrete motion tokens and unifies them with text tokens, treating motion as a specific language and achieves state-of-the-art on tasks like text-driven generation and motion captioning. The advantage is that powerful language-model techniques (prompt learning, large-scale pretraining) can be directly applied to motion.
    
        Models like MMM~\cite{pinyoanuntapong2024mmm} use a BERT-like masked-token objective. During training, a fraction of tokens is masked and the model learns to reconstruct them given text and context. MMM’s motion tokenizer and masked motion model jointly learn to regenerate the original token sequence. Such masked modeling can support flexible generation and editing (e.g. inpainting by inserting [MASK] tokens), at the cost of a more complex training scheme than simple autoregression. \\
    
        More recently, methods have introduced diffusion processes in the discrete token space. For example, MoDDM~\cite{Chemburkar_2023_BMVC} and the M2D2M model~\cite{chi2024m2d2m} use a VQ-VAE to obtain discrete motion codes and then learn a denoising diffusion probabilistic model (DDPM) over token sequences. MoDDM explicitly corrupts VQ-VAE codes in the latent space and learns to denoise them back to valid tokens. M2D2M extends this idea to multi-action generation, introducing a dynamic transition probability in the discrete diffusion process to ensure smooth transitions between different actions. These hybrid discrete-diffusion models aim to combine the structured token representation with the probabilistic richness of diffusion. \\
    
        \paragraph{} Advantages of discrete representations include the ability to leverage NLP architectures and training techniques. The tokenized form naturally handles variable-length motion as a sequence of symbols, and text and motion can share similar modeling frameworks. Discrete tokens also compress motion data (e.g. MMM’s large codebook yields high fidelity)~\cite{pinyoanuntapong2024mmm}. In practice, Zhang et al. report that T2M-GPT~\cite{zhang2023t2m} achieves a much lower FID than MotionDiffuse (0.116 vs 0.630) on HumanML3D, suggesting that discrete pipelines can produce very high-quality motions given sufficient data and codebook capacity. Nevertheless, Training VQ-VAEs can be challenging: large codebooks (8192 codes) are needed to avoid “information loss” but can suffer codebook collapse (many codes unused). MMM~\cite{pinyoanuntapong2024mmm} combats this with factorized codebooks and EMA resets, but such tricks complicate training. Discretization also inherently loses precision – the finest motion nuances can be lost unless the codebook is extremely large. Indeed, Zhang et al.~\cite{zhang2023t2m} note that T2M-GPT’s performance is ultimately limited by HumanML3D’s dataset size, implying that without more data (and codes) the token model cannot capture all motion variation. Finally, inference in discrete models can be slow if using autoregression, and streaming or dynamic generation is hindered by non-causal token sequences. MotionStreamer~\cite{xiao2025motionstreamer} explicitly points out that prior GPT-based methods cannot easily do online generation because tokens must be decoded in sequence.\\

    \subsubsection{Continuous Representations}
        \paragraph{} Continuous latent spaces represent motion as vectors in a high-dimensional, continuous space. These are often modeled using VAEs, transformers, or denoising diffusion probabilistic models (DDPMs). This approach enables a more fluid and expressive modeling of motion dynamics, including fine-grained articulations and smooth transitions. In practice, most continuous approaches use diffusion or autoencoder-based generators to sample motion trajectories or latent embeddings. \\

        \paragraph{} MotionDiffuse~\cite{zhang2022motiondiffuse} is a pioneer of continuous diffusion for text-to-motion. It treats motion as a high-dimensional sequence, injects noise at multiple levels, and iteratively denoises conditioned on text. The model emphasizes probabilistic mapping (stochasticity) and realistic synthesis, claiming superior diversity and fidelity. Similarly, FLAME~\cite{kim2023flame} and MoFusion~\cite{dabral2023mofusion} integrate diffusion into motion synthesis. FLAME uses a transformer-based diffusion model that can edit motions frame- or joint-wise without retraining. MoFusion focuses on long and physically plausible motions from varied conditions (text, music) and even incorporates kinematic loss terms to enforce realism. All these methods operate on continuous pose representations or learned continuous latents, without discretizing to tokens.\\

        Some methods learn a continuous latent space via an autoencoder and run diffusion in that latent space. For example, SALAD~\cite{hong2025salad} introduces a skeleton-aware latent diffusion model where a VAE compresses motion into a latent; the diffusion denoiser explicitly models joint-frame-word relationships, yielding strong text-motion alignment and zero-shot editing via cross-attention. MoLA~\cite{uchida2024mola} similarly integrates a VAE and latent diffusion, enhanced by adversarial training, to achieve “high-quality and fast” text-to-motion generation. These latent-diffusion hybrids avoid the expensive dimensionality of raw diffusion and allow leveraging VAE priors, bridging the discrete/continuous gap to some extent.

        Newer models like MotionStreamer~\cite{xiao2025motionstreamer} combine autoregressive and diffusion elements in a continuous latent space. MotionStreamer uses a causal autoencoder to produce continuous motion latents and a diffusion “head” to predict next-latent tokens in a streaming fashion. It explicitly contrasts its continuous approach with prior discrete-token methods, noting that continuous latents “avoid information loss of discrete tokens” and error accumulation. \\

        \paragraph{} Advantages of continuous methods center on expressiveness and fidelity. Because they operate in a real-valued space, they can capture subtle variations without quantization error. Diffusion models are known to model complex distributions faithfully; MotionDiffuse’s probabilistic denoising yields diverse samples and fine-grained text adherence~\cite{zhang2022motiondiffuse}. Editing and conditioning can be more natural: FLAME~\cite{kim2023flame} edits parts of motion directly in the continuous sequence, and latent-diffusion methods like SALAD~\cite{hong2025salad} can leverage continuous attention weights for zero-shot edits. Training-wise, diffusion models have a simple, stable denoising objective (often MSE-based) that avoids issues like codebook collapse. MoFusion~\cite{dabral2023mofusion} even integrates known motion priors (kinematic losses) into training, something less straightforward in a tokenized pipeline.

        However, Iterative diffusion requires dozens or hundreds of forward passes per generated sequence, which is slower than a single-pass autoregressive decode. Models may also be sensitive to the choice of representation: e.g., FLAME~\cite{kim2023flame} had to design a special transformer to handle variable-length continuous sequences. Continuous models may require vast compute to train (diffusion UNets can be large) and may struggle with very long sequences. Furthermore, alignment with language can be less direct: discrete tokens have an inherent language-like structure that diffusion lacks. As a result, some continuous models incorporate expensive cross-attention with text (FLAME~\cite{kim2023flame}, SALAD~\cite{hong2025salad}) or extra signals to improve alignment. Finally, while diffusion covers data modes well, it can still miss rare motions without careful training; continuous latent methods like MoLA~\cite{uchida2024mola} use adversarial losses to encourage realism beyond the averaged diffusion prior.

    \subsubsection{Discussion}
        There is a growing interest in hybrid paradigms that blend discrete and continuous representations. One class is latent diffusion models (LDMs): these first compress motion into a lower-dimensional latent via an encoder, then run a diffusion process in that latent. By operating in a compact continuous space, LDMs aim for the best of both worlds. SALAD and MoLA (above) are examples: they use learned latents but retain continuous stochastic generation. Another hybrid path is diffusion on discrete tokens, as in MoDDM~\cite{Chemburkar_2023_BMVC}and M2D2M~\cite{chi2024m2d2m}. These methods keep the VQ-VAE tokenization but apply a diffusion-style prior on the token sequence. This can provide richer mixing of modes (e.g. the dynamic transition in M2D2M that blends action tokens smoothly) while still utilizing a discrete structured vocabulary. MotionStreamer~\cite{xiao2025motionstreamer} similarly blends an autoregressive context encoder with a diffusion decoder for continuous latents. 

        Such hybrids attempt to reconcile discrete and continuous advantages. For instance, LDMs benefit from continuous smoothness and reduced dimensionality, while discrete diffusion retains interpretable tokens with richer uncertainty modeling. MoLA’s combination of a VAE and diffusion (plus GAN loss)~\cite{uchida2024mola} yields faster sampling and high quality. M2D2M~\cite{chi2024m2d2m} shows that discrete diffusion can extend single-action models to multi-action synthesis with coherent transitions. In general, these works suggest that mixing paradigms can mitigate each approach’s weaknesses: using a VAE latent (continuous) addresses token quantization issues, while using a tokenized interface can leverage powerful language priors. 
        Table~\ref{tab:motion-space-cat} provides a comparison between discrete and continuous motion representations, highlighting their respective advantages, challenges, and implications for modeling and generation quality.

    \paragraph{} In summary, discrete and continuous latent paradigms each have merits. Discrete models (T2M-GPT~\cite{zhang2023t2m}, MotionGPT~\cite{jiang2023motiongpt}, MMM~\cite{pinyoanuntapong2024mmm}) excel at leveraging language-model inductive biases and have shown surprisingly high quality when well-trained. Continuous methods (MotionDiffuse~\cite{zhang2022motiondiffuse}, FLAME~\cite{kim2023flame}, MoFusion~\cite{dabral2023mofusion}) harness diffusion’s expressiveness to achieve vivid and controllable motion with probabilistic diversity. Hybrid approaches are gaining traction as they blend these strengths. Moving forward, we anticipate more unification: for instance, latent diffusion models that combine efficient representation with powerful text guidance, and token-diffusion models that allow fine-grained multi-action control (e.g. M2D2M~\cite{chi2024m2d2m}). Other directions include faster samplers for diffusion, larger and more diverse motion-language pretraining (as exemplified by MotionGPT~\cite{jiang2023motiongpt}), and methods that better exploit the structure of motion (like skeleton-awareness in SALAD~\cite{hong2025salad}). Ultimately, a flexible framework that can seamlessly bridge discrete token semantics and continuous motion dynamics may provide the best of both worlds for text-to-motion generation.

\section{Datasets and Evaluation Metrics}

\subsection{Language-Motion Datasets}

    \paragraph{} Text-to-motion generation models rely heavily on the availability of high-quality, diverse, and well-annotated datasets that pair human motion sequences with corresponding natural language descriptions. As the field has evolved, so too has the landscape of datasets, spanning various modalities such as action-conditioned, speech-conditioned, and scene-aware motion generation, across both 2D and 3D domains. In this section, we focus on datasets designed specifically for text-driven human motion generation—resources that form the cornerstone of training, benchmarking, and advancing model performance.

    \paragraph{} Three primary datasets have emerged as benchmarks in the field: KIT~\cite{Plappert2016}, BABEL~\cite{BABEL:CVPR:2021}, and HumanML3D~\cite{Guo_2022_CVPR}. The KIT Motion-Language dataset was the first of its kind to align human motion capture sequences with natural language. It provides 6,278 sentence-level descriptions for 3,911 unique motions performed by 111 different subjects. These motions were aggregated from various motion capture databases~\cite{mandery2015kit,cmu_dataset} and unified under the Master Motor Map (MMM) framework~\cite{terlemez2014master}, offering a coherent structure for joint representation and processing.

    \paragraph{} BABEL significantly expanded the scope and granularity of language-motion alignment by annotating approximately 43 hours of motion from the AMASS dataset~\cite{mahmood2019amass}. This dataset distinguishes itself through a dual-level annotation scheme, consisting of sequence-level labels that describe the overarching activity, and frame-level labels that detail temporally localized actions. BABEL includes over 28,000 sequence labels and 63,000 frame labels, covering more than 250 action categories, making it one of the most semantically rich datasets available.

    \paragraph{} HumanML3D~\cite{Guo_2022_CVPR} further advances the field by offering a high-resolution pairing of language and motion data. It compiles 14,616 motion sequences—sourced from AMASS and HumanAct12~\cite{guo2020action2motion}—along with 44,970 human-annotated textual descriptions comprising 5,371 distinct words. The dataset encompasses approximately 28.6 hours of motion and provides an ideal testbed for evaluating models on fine-grained linguistic grounding and motion expressiveness.

    \paragraph{} Recent efforts have aimed at expanding the breadth and complexity of such datasets. HuMMan-MoGen~\cite{zhang2023finemogen} builds upon the HuMMan dataset~\cite{cai2022humman} by manually annotating 2,968 video clips with 112,112 textual descriptions, resulting in 6,264 motion sequences spanning 179 subjects. While smaller in scope, its curated nature and video-level alignment offer opportunities for bridging video-to-motion synthesis and generation.

    \paragraph{} The most comprehensive dataset to date is Motion-X~\cite{lin2024motion}, which introduces a novel focus on full-body human motion, including intricate details such as hand gestures and facial movements, via SMPL-X body modeling~\cite{pavlakos2019expressive}. It includes 81,084 motion sequences, totaling 144.2 hours of motion, along with over 15.6 million 3D pose annotations and synchronized RGB video frames. Motion-X thus enables multi-modal learning and supports advanced tasks like 3D mesh recovery and fine-grained gesture synthesis.

    \paragraph{} Table~\ref{tab:language-motion-datasets} provides a comparative summary of these datasets, highlighting their scale, duration, annotation types, and anatomical coverage.

\begin{table*}[ht]
    \centering
    \caption{Comparison of Language-Motion Datasets}
    \label{tab:language-motion-datasets}
    \begin{center} 
    \begin{tabular}{lW{c}{1.5cm}W{c}{1.65cm}W{c}{4cm}W{c}{3cm}}
        \toprule
        \textbf{Dataset} & \textbf{\# Sequences} & \textbf{Total Hours} & \textbf{Annotations} & \textbf{Body Coverage} \\
        \midrule
        KIT~\cite{Plappert2016} & 3,911 & $\sim$6 & Sentence Descriptions & Upper/Lower Body \\
        BABEL~\cite{BABEL:CVPR:2021} & $\sim$28,000+ & $\sim$43 & Seq + Frame Labels & Full Body \\
        HumanML3D~\cite{Guo_2022_CVPR} & 14,616 & 28.6 & Sentences (44,970) & Full Body \\
        HuMMan-MoGen~\cite{zhang2023finemogen} & 6,264 & - & 112,112 Descriptions & Full Body \\
        Motion-X~\cite{lin2024motion} & 81,084 & 144.2 & Seq Descriptions + 3D Poses & Whole Body (incl. hands, face) \\
        \bottomrule
    \end{tabular}
    \end{center}
\end{table*}

\subsection{Evaluation Metrics}

    \paragraph{} Evaluating the quality and effectiveness of generated motion remains a core challenge in the field of text-to-motion generation. While human visual assessment offers valuable insights, it is inherently subjective and resource-intensive. Consequently, a suite of quantitative metrics has been proposed to assess different facets of model performance—namely fidelity, diversity, and text-motion alignment.

    \paragraph{} The five most commonly adopted metrics in the literature are: Fréchet Inception Distance (FID), R-Precision, Diversity, Multi-Modality (MM), and Multi-Modal Distance (MM-Dist). Each of these targets a distinct aspect of generation quality.

    FID, originally developed for image generation, has been adapted to the motion domain by embedding motion sequences into a learned feature space (typically via a pre-trained motion encoder) and computing the Fréchet distance between the distributions of real and generated motions. Lower FID scores indicate better alignment and realism.

    R-Precision quantifies semantic consistency by evaluating how accurately the generated motion can be matched back to its corresponding textual description in a retrieval setting. It is computed as the top-$k$ retrieval accuracy and reflects the faithfulness of generated motions to their textual prompts.

    Diversity and Multi-Modality metrics assess the range of outputs the model can produce. Diversity measures the variance among different generated motions for different prompts, while Multi-Modality focuses on variations in generated motions for the same prompt—capturing the model's capacity for one-to-many generation.

    MM-Dist serves as a text-motion alignment metric. It computes the average distance between the embedding of a text description and its associated motion feature, helping to quantify how closely the model has captured the semantics of the input text. \\

    \paragraph{} Despite the usefulness of these quantitative metrics, they come with limitations. Most notably, they require access to ground-truth data and often fail to capture fine-grained perceptual aspects of motion such as style, subtle intent, or emotional expressiveness. Furthermore, they might not generalize well across different motion lengths, body parts, or actions not seen during training. Therefore, qualitative evaluation remains a crucial complement to quantitative metrics. Visual inspection, user studies, and expert judgments continue to play an important role in validating motion realism, fluidity, and semantic alignment in open-ended scenarios.

    \paragraph{} As the field progresses, future benchmarks may incorporate interactive and task-based evaluations, such as in-the-loop reinforcement tasks or user preference rankings, to more robustly assess model performance in real-world applications.

\section{Current Challenges and Limitations}\label{sec:challenges}

    \paragraph{} Despite the remarkable progress in text-driven motion generation, the field still faces several critical challenges that hinder its development and scalability. In this section, we highlight three principal limitations: the scarcity of data, the lack of fine-grained motion details, and the limitations in evaluation methods.
    
    \subsection{Lack of Datasets}
        \paragraph{} Unlike fields such as text, image, or video generation, motion generation suffers from a significant lack of large-scale, high-quality, and semantically rich datasets. Most available datasets, such as HumanML3D~\cite{Guo_2022_CVPR} or Motion-X~\cite{lin2024motion}, contain only a few tens of thousands of motion samples, making them limited in diversity and coverage. This scarcity originates from the fact that 3D motion data, such as meshes or skeletons, are not naturally abundant and typically require costly preprocessing, e.g. motion capture or video-to-motion estimation using pretrained models. These procedures are computationally expensive and often introduce noise and artifacts. Consequently, the ability to train highly generalizable models capable of synthesizing arbitrary activities from text remains an open challenge. In contrast, image and text models benefit from access to billions of training samples, allowing them to learn rich semantic priors.
    
        \paragraph{} Moreover, while motion data can be implicitly found in videos, converting these videos into accurate 3D representations is non-trivial. Thus, a promising direction for the future involves training models directly on large-scale video datasets, where motion is inherently encoded~\cite{feng2024chatpose,lin2024chathuman}. This paradigm shift could unlock the scalability needed for achieving general-purpose motion generation systems.
    
    \subsection{Fine-Grained Motion Details}
        \paragraph{} Another critical limitation lies in the lack of fine-grained motion details, such as hand gestures and facial expressions. Most existing methods focus on full-body skeletal motion, neglecting subtler motions essential for activities like playing piano or expressive communication. For example, generating motions where only fingers move while the rest of the body remains stationary remains an unsolved problem. This limitation is partly due to the absence of integrated datasets that contain synchronized face, finger, and body motion annotations. Recent efforts have begun addressing these issues by treating hand and face synthesis as independent subproblems, but holistic and temporally consistent solutions remain scarce.

    \subsection{Evaluation Ambiguity}
        \paragraph{} Evaluating generated motions is inherently complex. While several quantitative metrics have been proposed—such as FID, diversity scores, and motion realism indices~\cite{ismail2024establishing}—none fully capture all aspects of motion quality, especially for nuanced behaviors. Most metrics require ground truth data for comparison, which is often unavailable, especially for open-ended or user-driven prompts. Furthermore, these metrics struggle to assess temporal coherence, naturalness, and semantic alignment with the text. As a result, qualitative evaluation remains indispensable. Human studies and perceptual assessments continue to play a crucial complementary role, highlighting the need for developing standardized, interpretable, and multi-dimensional evaluation frameworks. \\

    \paragraph{} In addition to the previous challenges, two further challenges persist in text-driven motion generation: the complexity of motion representation and the difficulty in preserving temporal coherence. Representing human motion in a way that is both expressive and learnable remains unresolved. Common formats such as joint coordinates, axis-angle rotations, and parametric body models like SMPL all entail trade-offs between realism, dimensionality, and modeling ease. Discrete motion representations offer simplicity, making learning more manageable, but they often fall short in capturing the dynamic changes in movement over time. In contrast, continuous representations maintain a higher level of detail and temporal accuracy, though they come with the challenge of increased modeling complexity. Another issue many models face, especially autoregressive ones, is maintaining consistency over time. These models often struggle with long sequences, leading to issues like foot sliding, jitter, or unnatural motion discontinuities. While diffusion-based methods have made strides in improving long-term consistency, they tend to be computationally expensive and require sophisticated techniques to align the generated motion with the input text across the entire sequence. These challenges highlight the need for better motion representations and temporal modeling strategies that strike the right balance between realism, efficiency, and interpretability.

\section{Emerging Trends and Future Directions}\label{sec:emerging-trends}
    
    \paragraph{} With rapid advances in generative models and increasing interest from the research community, several promising directions are emerging in the field of text-driven motion generation. We outline three central themes expected to shape future work: large-scale motion models, motion editing and controllability, and scene-aware motion generation.
    
    \subsection{Large-Scale Motion Models}
        \paragraph{} The integration of large-scale pretraining paradigms, inspired by language and vision models, is beginning to gain traction in motion generation. For instance, MotionGPT~\cite{zhang2024motiongpt} explores the use of pretrained LLMs to translate text prompts into motion representations. Similarly, efforts such as PoseScript~\cite{delmas2022posescript} and MotionLLM~\cite{chen2024motionllm} propose question-answering-style datasets to better understand fine-grained motion-text relationships. A recent work, QuoVM~\cite{Wang2024QuoVM}, directly addresses the viability of large-scale motion models trained on extensive datasets, emphasizing the importance of scaling both model size and data volume.
    
        \paragraph{} Future research is likely to pivot towards training generalist motion models directly from video datasets, where motion, context, and audio cues can be naturally embedded and learned in a multimodal fashion. Such models promise significantly greater coverage, semantic understanding, and diversity in generated motions.
    
    \subsection{Motion Editing and Controllability}
        \paragraph{} Controllability refers to the ability to precisely guide or manipulate generated motions based on user-defined constraints or instructions. This includes tasks such as motion retargeting, style transfer, in-betweening, and temporal editing~\cite{karunratanakul2023guided}. While current models allow for some degree of style blending and sequential action synthesis, editing motions in a fine-grained, user-friendly manner remains an open challenge.
    
        \paragraph{} Future systems are expected to incorporate more flexible and expressive control mechanisms—potentially using natural language, spatial constraints, or keyframe supervision—to allow for interactive motion editing without compromising realism or temporal consistency.
    
    \subsection{Scene-Aware and Interactive Motion Generation}
        \paragraph{} Most existing models generate motions in isolation, neglecting the spatial and physical constraints of real-world environments. Scene-aware motion generation, which considers interactions between the human and the surrounding environment (e.g., obstacles, objects, other people), is vital for realistic and deployable systems~\cite{wang2021synthesizing,zhao2023synthesizing}.
    
        \paragraph{} This includes human-scene interaction (e.g., avoiding a table while walking or sitting on a chair) and human-human interaction (e.g., two people dancing or shaking hands). Such contexts demand awareness of physical feasibility, mutual coordination, and spatial awareness. Future research is likely to focus on integrating scene perception, object affordances, and physics simulation into motion synthesis pipelines to produce more grounded and realistic behaviors.

    \paragraph{} Beyond the key directions discussed, several additional avenues merit attention in future research. One promising path is multimodal integration, where audio, visual, and linguistic signals are fused to produce more context-aware and expressive motions. For instance, aligning speech prosody with gesture dynamics or adapting motions based on scene context from images or videos could lead to significantly richer human behavior synthesis. Additionally, moving beyond skeletal representations to full-body mesh models such as SMPL-X~\cite{pavlakos2019expressive} would allow capturing subtle cues like facial expressions and finger articulations, expanding applicability to emotionally nuanced and interaction-heavy scenarios like virtual reality, gaming, and digital avatars. These enhancements, though technically demanding, are essential for creating more immersive and human-like motion generation systems.

\section{Conclusion}\label{sec:conclusion}
    \paragraph{} This review presents a comprehensive examination of both traditional and modern approaches in the field of human motion generation, with a particular emphasis on text-driven methods. Traditional methods primarily treated motion synthesis as a prediction problem, relying on initial pose sequences and, in some cases, conditioning on action labels encoded as one-hot vectors. While foundational, these methods are inherently constrained in flexibility, diversity, and realism.

    \paragraph{} In contrast, modern text-to-motion generation frameworks capitalize on the rapid progress in natural language processing and generative modeling. These techniques enable direct synthesis of human motion from text prompts—ranging from simple action labels to rich natural language descriptions—without requiring prior motion context. We categorized state-of-the-art models from two complementary perspectives: (i) \textbf{architecture-based}, including VAE-based, diffusion-based, and hybrid approaches, and (ii) \textbf{motion-space-based}, distinguishing between discrete and continuous motion generation paradigms.

    \paragraph{} Beyond model categorization, this paper explores the critical datasets and evaluation metrics shaping the development of the field. We further highlight current challenges including (i) the scarcity of large, diverse, and richly annotated language-motion datasets, (ii) the complexities of motion representation and synthesis fidelity, and (iii) the limitations of autoregressive models, particularly in comparison to the superior generative quality offered by diffusion-based frameworks.

    \paragraph{} Looking ahead, we foresee promising research directions in incorporating multi-modal data (e.g., vision and audio), refining motion realism through mesh-level synthesis, enhancing control through prompt engineering and fine-grained text understanding, and designing more efficient, scalable generative architectures. By consolidating existing work, clarifying key issues, and identifying emerging trends, we aim for this survey to serve as a guiding resource for researchers and practitioners driving the future of language-conditioned motion generation.





\bibliographystyle{named}




\end{document}